 \documentclass[letterpaper, 10 pt, conference]{ieeeconf}

\IEEEoverridecommandlockouts                              

\usepackage{amsmath} 
\usepackage{bm} 
\usepackage{tikz}
\usetikzlibrary{arrows.meta,calc,positioning,fit}
\usepackage{amssymb}  
\usepackage{subcaption}
\usepackage{float}
\usepackage{subcaption}
\usepackage{multirow}
\usepackage{booktabs}
\usepackage{algorithm}
\usepackage[noend]{algpseudocode}
\usepackage{svg}

\title{\LARGE \bf
Neural-Surrogate Co-Design of Aerodynamic Shape and Control
}

\author{Daniele Affinita, Ming Xu, Beno\^{i}t Gherardi, Pascal Fua%
\thanks{All authors are with \'{E}cole Polytechnique F\'{e}d\'{e}rale de Lausanne (EPFL), Lausanne, Switzerland
        \{daniele.affinita, mingda.xu, benoit.gherardi, pascal.fua\}@epfl.ch}}

\newif\ifdraft
\draftfalse
\drafttrue

\ifdraft
\newcommand{\PF}[1]{{\color{red}{\bf PF: #1}}}

\newcommand{\MX}[1]{{\color{blue}{\bf MX: #1}}}

\newcommand{\DA}[1]{{\color{teal}\textbf{DA: #1}}}

\else
\newcommand{\PF}[1]{}

\newcommand{\MX}[1]{}

\newcommand{\DA}[1]{}

\fi

\newcommand{\parag}[1]{\noindent{\bf #1}}

\renewcommand{\eqref}[1]{Eq.~(\ref{#1})}

\begin{document}

\maketitle
\thispagestyle{empty}
\pagestyle{empty}


\begin{abstract}

Control co-design jointly optimizes a robot's physical design and its controller. Existing techniques are limited in the number of model parameters and the complexity of the physical models they can handle. In this paper, we propose a gradient-based bilevel formulation that replaces the dynamics models of earlier techniques with a differentiable neural surrogate and differentiates through the trajectory optimizer using a structure-exploiting approach. We validate the framework on fixed-wing gliders, using NeuralFoil as the aerodynamic surrogate. The tasks are maximum-distance gliding, minimum-distance landing, and perching. Experiments on 2D airfoils and 3D wings show that our method runs about five times faster than a state-of-the-art evolutionary baseline. It also outperforms pipelines that optimize aerodynamic proxy metrics.

\end{abstract}

\section{Introduction}

The design of robotic systems is inherently coupled with the controllers that operate them. Thus, it makes sense to jointly optimize the physical design and the controller, an approach often referred to as \textit{Control Co-Design} (CCD)~\cite{GarciaSanz19,Herber19,allison2010combined,Martins13a,Bergonti24,Fernandez24}.
%
Existing CCD approaches can be broadly categorized into simultaneous, discrete, sampling-based, and gradient-based methods. The first two categories are best suited for relatively low-dimensional models~\cite{Herber19,Carlone19,Zhao22b,Wauters22,Bergonti24}. Gradient-based nested co-design addresses this scalability problem by solving the control problem in the inner loop and having the outer loop update the design using the gradient of the mission cost. However, the nested structure of CCD introduces major computational challenges that force the use of simplified physics-based models and restrict the applicability of the methods~\cite{Lee16b,Mackle24}. 

To address the computational challenge and broaden the applicability of the CCD, we retain the bilevel formulation of co-design~\cite{Lee16b,Mackle24} but replace the dynamics model with a differentiable neural surrogate and differentiate through the trajectory optimizer using a structure-exploiting approach~\cite{Xu23g}. Evaluating and differentiating the surrogate is cheaper than performing a full simulation. This helps but would not be enough by itself. A naive implementation would still require querying and computing first and second-order derivatives through the full network at every time step of every solver iteration, which would be intractable for networks with millions of parameters. We solve this by exploiting the structure of the bilevel problem: For a fixed design, the surrogate needs only to be queried a few times per outer iteration, and the inner loop can then operate on an easy-to-differentiate approximation of it. We also account for another weakness of surrogate models:  Their predictions are typically valid only in restricted regions of the design space, and the optimizer can easily take the design out of that part of the space, resulting in degenerate shapes and unrealistic predictions~\cite{Seiler26a}. We use the surrogate's confidence in its own predictions to prevent this.


\begin{figure}
    \centering
    \includegraphics[width=\linewidth]{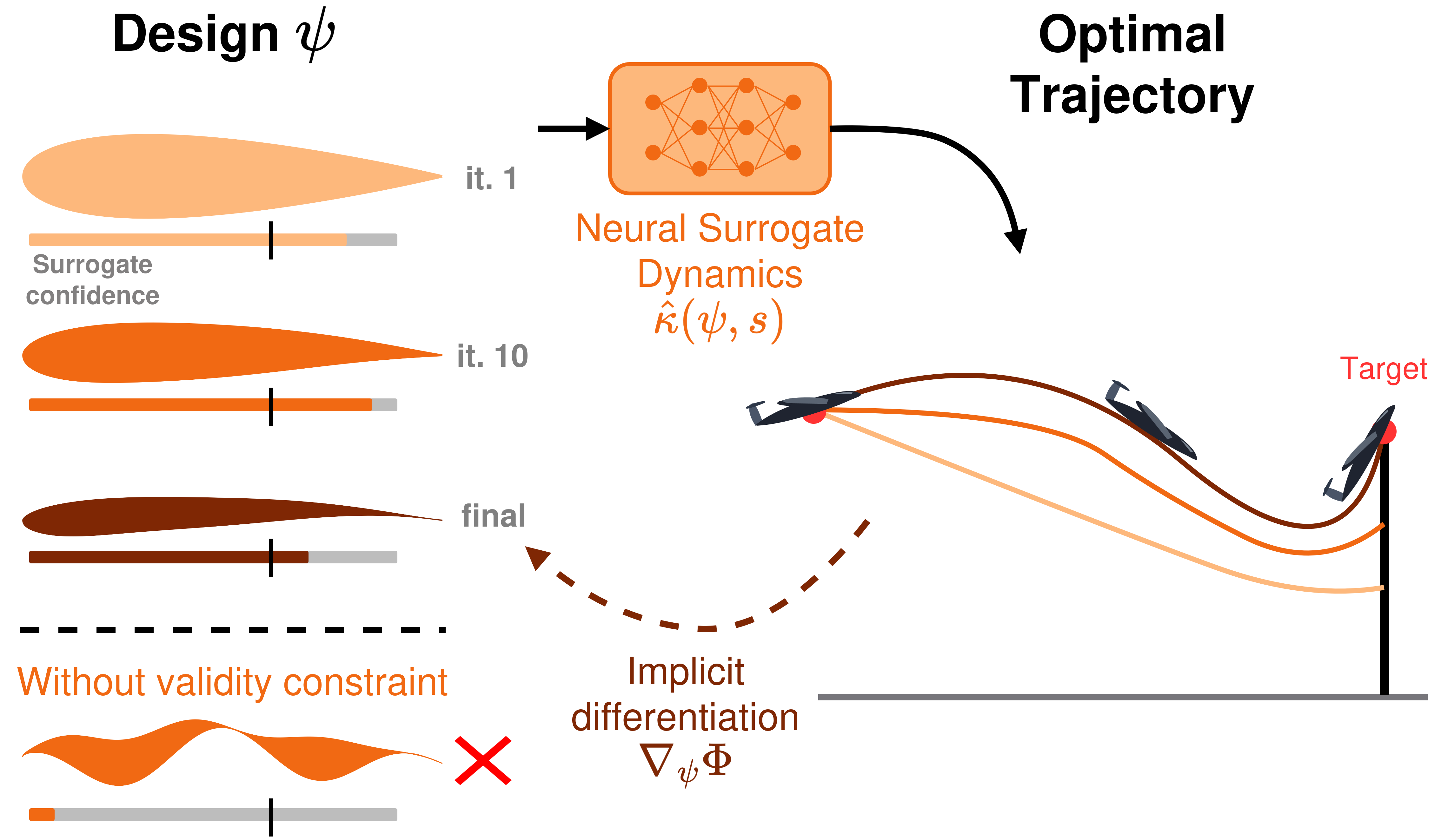}
    \vspace{-5mm}
    \caption{\textbf{Co-design}. We jointly optimize the airfoil shape (left, colored by iteration) and the controls. The shape determines the trajectories (right) through a neural surrogate, and the gradient flows back by implicit differentiation. Without the validity constraint (bottom), the optimizer drives the shape to where the surrogate is unreliable.}
    \vspace{-6mm}
    \label{fig:teaser}
\end{figure}

We validate our framework on fixed-wing gliders having to perform three increasingly complex tasks: Maximum-distance gliding, minimum-distance landing, and perching, as illustrated by  Fig.~\ref{fig:teaser}, using NeuralFoil~\cite{Sharpe25a} as the differentiable aerodynamic surrogate. This constitutes a challenging testbed for our method for several reasons. First, the dynamics equations are only partially available in closed form, as aerodynamic coefficients must be estimated through expensive computational fluid dynamics (CFD) simulations. Second, because gliders are unpowered, their performance is determined solely by their aerodynamic design and control strategy, without thrust compensation for suboptimal designs. Third, there are several potential flight regimes, making the physics truly challenging. This is especially true of the perching task. It has been demonstrated before~\cite{Moore14,Waldock18} but never in a co-design setting where we optimize for both control inputs and wing shape. One reason is that it requires operating in the post-stall regime, which is excluded by previous gradient-based co-design approaches. Finally, these tasks require different aerodynamic solutions, making it necessary to optimize the design directly from the task objective rather than relying on manually selected aerodynamic proxy metrics.
 
Our main contribution is a generic gradient-based nested co-design framework in which the dynamics are modeled in terms of a neural surrogate, removing the restrictive assumptions of prior gradient-based pipelines~\cite{Lee16b,Mackle24}. It retains exact sensitivities throughout the optimal control problem, and its computational cost only grows linearly with the planning horizon. Experiments on 2D airfoil and 3D wing design show that our framework outperforms a state-of-the-art evolutionary co-design baseline by about five times in terms of computation time, as well as standard pipelines based on optimizing aerodynamic proxy metrics in terms of performance. 


\section{Related Work}
\label{sec:related}


%

In robotics, co-design is usually formulated over discrete component choices, such as motors, sensors, and actuators, or low-dimensional continuous parameterizations. These problems can be formulated as binary linear programs that maximize task performance~\cite{Carlone19}. Alternatively, evolutionary algorithms can jointly optimize design and control parameters \cite{Zhao22b, Bergonti24}. However, they require many expensive simulations and scale poorly with the dimensionality of the search space~\cite{Zhou24a}, making them unsuitable for high-dimensional continuous problems. Hybrid discrete–continuous formulations have also been proposed to optimize topology choices and continuous parameters simultaneously~\cite{Zhao22b}, but they remain suited for structured design spaces rather than direct optimization of complex geometries. Sampling-based optimization methods, such as Bayesian Optimization, provide an alternative for black-box problems, but their sample efficiency deteriorates as the number of design variables increases~\cite{Malu21a}. Consequently, they are rarely used for direct optimization of complex shapes in high-dimensional spaces. 

\begin{figure*}[t]
\centering
\vspace{4pt}

\definecolor{ink}{HTML}{262626}\definecolor{sub}{HTML}{6E6E6E}\definecolor{line}{HTML}{9A9A9A}
\definecolor{palBrown}{HTML}{7F2704}\definecolor{palOrange}{HTML}{EC620F}\definecolor{palLight}{HTML}{FDB576}
\colorlet{netFill}{palLight!40!white}\colorlet{netLine}{palOrange}
\colorlet{gradc}{palBrown}\colorlet{valc}{palOrange}
\begin{tikzpicture}[x=1cm, y=1cm, font=\small, node distance=0.6cm,
  blk/.style={draw=line, line width=0.5pt, fill=white, rounded corners=2pt,
              minimum height=1.15cm, align=center, inner sep=4pt, text=ink},
  fwd/.style={-{Latex[length=4pt,width=3pt]}, line width=0.6pt, draw=ink},
  grad/.style={-{Latex[length=4pt,width=3pt]}, line width=0.7pt, draw=gradc,
               dash pattern=on 3pt off 2pt, rounded corners=3pt},
  val/.style={-{Latex[length=4pt,width=3pt]}, line width=0.7pt, draw=valc,
              dash pattern=on 1pt off 1.5pt, rounded corners=3pt},
  lab/.style={font=\footnotesize, inner sep=2pt},
]
\node[blk, minimum width=1.9cm] (shape) {Shape $\psi$\\[-1pt] {\scriptsize\color{sub} Updated using}\\[-3pt]
  {\scriptsize\color{sub} augmented Lagrangian}};
\node[blk, fill=netFill, draw=netLine, right=of shape] (net)
  {Neural surrogate\\[-1pt]
   {\scriptsize\color{sub} on the flight envelope $\mathcal{E}$}};
\node[blk, right=of net] (fit)
  {Polynomial fit\\[-1pt] {\scriptsize\color{sub} Chebyshev reduction}};
\node[blk, minimum width=3.4cm, right=0.75cm of fit, draw=none, fill=none, text opacity=0] (traj)
  {Trajectory optimization\\[-1pt] {\scriptsize one per task}};
\node[blk, minimum width=3.4cm, fill=white] (tB) at ($(traj)+(0.16,0.16)$)
  {\phantom{Trajectory optimization}\\[-1pt] {\scriptsize\phantom{one per task}}};
\node[blk, minimum width=3.4cm, fill=white] (tA) at ($(traj)+(0.08,0.08)$)
  {\phantom{Trajectory optimization}\\[-1pt] {\scriptsize\phantom{one per task}}};
\node[blk, minimum width=3.4cm] at (traj)
  {Trajectory optimization\\[-1pt] {\scriptsize\color{sub} one per task}};
\node[blk, minimum width=2.4cm, right=1.0cm of tB.east |- traj, anchor=west] (cost) {Mission cost $\Phi$};
\node[fit=(traj)(tB), draw=line, line width=0.5pt, dash pattern=on 3pt off 2pt,
      rounded corners=4pt, inner sep=0.14cm] (inner) {};
\node[lab, text=sub, font=\scriptsize\itshape, anchor=south east] at (inner.north east)
  {inner loop};
\draw[fwd] (net) -- node[lab, above] {$\hat\kappa$} (fit);
\draw[fwd] (shape) -- (net);
\draw[fwd] (net) -- (fit);
\draw[fwd] (fit) -- node[lab, above] {$w$} (traj);
\draw[fwd] (tB.east |- traj) -- (cost);
\node[lab, anchor=south, inner sep=1pt] at ($(inner.east |- traj)!0.5!(cost.west)+(0,0.04)$) {$x^*\!, u^*$};
\coordinate (vS) at ($(traj.north)+(-0.9,0.16)$);
\draw[val] (vS) -- ++(0,0.3) -| (net.north);
\node[lab, text=valc, anchor=south] at ($(vS)!0.5!(net.north |- vS)+(0,0.3)$)
  {surrogate confidence at the operating points};
\coordinate (gL) at ($(shape.south)+(0,-0.5)$);
\draw[grad] (cost.south) |- (gL) -- (shape.south);
\node[lab, text=gradc, anchor=north] at ($(fit.south |- gL)!0.5!(traj.south |- gL)$)
  {design gradient $\nabla_\psi \Phi$ by implicit differentiation};
\end{tikzpicture}
\caption{\textbf{Overview.} The neural surrogate is evaluated on the flight envelope once per
design update and replaced by a polynomial, which the trajectory optimizations, one per task and
launch state, use instead of the network. The design gradient and the validity constraint are
detailed in Secs.~\ref{sec:polynomial} and~\ref{sec:confidence}.}
\label{fig:pipeline}
\end{figure*}

Gradient-based nested co-design makes it possible to use more sophisticated models by solving the control problem in the inner loop and having the outer loop update the design using the gradient of the mission cost~\cite{Lee16b,Mackle24}. However, the nested structure of CCD introduces significant computational challenges. At each iteration of the outer design optimization, the inner optimal control problem must be solved for the current design, requiring repeated evaluations of the system dynamics. When these dynamics are obtained from expensive physical models, such as high-fidelity aerodynamic or structural simulations, the computational cost can quickly become prohibitive. Moreover, gradient-based approaches require computing sensitivities of the optimal objective with respect to the design variables, which involves differentiating through the trajectory optimization problem. Efficient sensitivity computation is therefore essential to make nested CCD scalable to complex dynamical systems. As a result, existing gradient-based CCD methods are limited to settings where the dynamics can be evaluated and differentiated efficiently. For example, in~\cite{Mackle24} the flight envelope of the plane being designed is restricted, and the flow is assumed to be inviscid to limit the number of CFD calls, which confines the method to hypersonic flying regimes. Similarly, the approach of~\cite{Lee16b} relies on a linear policy and finite-difference sensitivities, which limits its maneuvers and its scaling with the design dimension.

\section{Methodology}
\label{sec:method}

We formulate the problem in Sec.~\ref{sec:formulation}. To make the approach computationally efficient, we built it on three ideas. First, we keep the equations of motion analytic and learn only the few design-dependent coefficients that enter them (Sec.~\ref{sec:neural}). Second, we exploit the nested structure of co-design: while optimizing a trajectory, the design remains fixed, so the learned model depends only on a low-dimensional operating point. We approximate this dependence with a polynomial after each design update. The trajectory optimizer therefore never evaluates the network, and its computational cost no longer depends on the network size (Sec.~\ref{sec:neural}). Third, since a learned model is only reliable near its training data, we constrain the design to regions where the model remains trustworthy across all conditions encountered during the maneuvers (Sec.~\ref{sec:confidence}). Finally, we combine these elements into a gradient-based optimization of the design (Sec.~\ref{sec:algorithm}). Fig.~\ref{fig:pipeline} shows the full pipeline.

\subsection{Problem Formulation}
\label{sec:formulation}

Let $\psi \in \mathbb{R}^d$ denote the design parameters of the robot shape, $x_t$ the system state at discrete time step $t$, and $u_t$ the control input. For a given design $\psi$, a finite-horizon optimal control algorithm returns the optimal controls $u^*_{1:N-1}(\psi)$ and the resulting trajectories $x^*_{1:N}(\psi)$.
For a single task, the co-design problem is to find
\begin{equation}
\begin{aligned}
\min_{\psi}  \quad & \Phi\big(\psi, x^*_{1:N}(\psi),
u^*_{1:N-1}(\psi)\big) \; , \\
\text{subject to}   \quad & h(\psi) = 0, \qquad g(\psi) \le 0,
\end{aligned}
\label{eq:upper}
\end{equation}
where $h$ and $g$ exclude degenerate or unmanufacturable designs. In the rest of the paper, we abbreviate the objective as $\Phi(\psi)$.

\parag{Multiple tasks.}
A robot is rarely designed for a single task. When a design must perform $J$ tasks, each with its own OCP and optimal cost $\Phi_j(\psi)$, minimizing a weighted sum of costs with different units requires arbitrary weight normalization and cannot recover non-convex regions of the Pareto front~\cite{Das97}. Instead, we minimize for one task and treat the others as requirements. We solve
\begin{equation}
\begin{aligned}
\min_{\psi}  \quad & \Phi_1(\psi)\;, \\
 \text{subject to}  \quad &
\Phi_j(\psi) \le \varepsilon_j,\; j = 2,\dots,J \\
  & g(\psi) \le 0, \quad h(\psi) = 0 .
\end{aligned}
\label{eq:multitask}
\end{equation}
This $\varepsilon$-constraint formulation \cite{Haimes71} requires no normalization, states each bound in its own task's units, and traces the full Pareto front, non-convex regions included, as $\varepsilon$ is swept. 

\parag{Optimal Trajectories.}
In both Eqs.~\ref{eq:upper} and~\ref{eq:multitask},  the optimal controls $u^*_{1:N-1}(\psi)$ are taken to be the solutions of 
\begin{equation}
\begin{aligned}
\min_{x_{1:N},\, u_{1:N-1}} \quad & \textstyle\sum_{t=1}^{N-1} \ell(x_t, u_t) + \ell_f(x_N)\; , \\
\text{subject to} \quad & x_1 = \hat{x}_1, \\
& c(x_t, u_t) = 0, \quad d(x_t, u_t) \le 0 \quad \forall t, \\
& x_{t+1} = f(x_t, u_t; \psi) \quad \forall t,
\end{aligned}
\label{eq:ocp}
\end{equation}
where $\ell(\cdot)$ is the stage cost, $\ell_f(\cdot)$ the terminal cost, $c(\cdot)$, $d(\cdot)$ represents the path constraints, and $f$ encodes the system dynamics. We use a gradient-based nonlinear programming solver, a common choice for constrained trajectory optimization \cite{Betts10}.

\parag{System Dynamics.}
In many robotic systems, the equations of motion are analytic, and the design affects only a few coefficients $\kappa$ in them. These coefficients are evaluated at the current operating condition $s_t = s(x_t, u_t)$ and we can reformulate the dynamics as
\begin{equation}
f(x_t, u_t; \psi) = \tilde{f}\big(x_t, u_t, \kappa(\psi, s_t)\big) .
\label{eq:structure}
\end{equation}
For a wing, $\kappa$ holds the lift, drag, and moment coefficients and $s_t$ the angle of attack and Reynolds number. Analogous decompositions hold for propellers (thrust and torque coefficients versus advance ratio), compliant elements (force versus deflection), and contacts (friction versus slip). In each case, $\kappa$ describes a local interaction, which depends on a few quantities rather than on the full state of the robot. The dimension of $s_t$ is thus set by the physics of the interaction and does not grow with the robot's degrees of freedom. We assume that $\kappa$ depends on the state only through $s_t$, meaning that the coefficients are quasi-steady. This structure has two consequences that our method relies on. First, only $\kappa$ needs to be learned since $\tilde{f}$ is exact and cheap to evaluate, so that learning $f$ as a whole would replace exact physics by an approximation for no gain. Second, $s_t$ has far fewer dimensions than $x_t$, which we exploit in Sec.~\ref{sec:polynomial}.

When $\kappa$ depends on complex physics, evaluating it can become expensive. This is particularly problematic for solving Eq.~\ref{eq:ocp}, where $\kappa$ and its derivatives must be evaluated at every time step and solver iteration. A learned model can make these evaluations much cheaper, but introduces a different problem: the optimizer may exploit errors in the learned model. Secs.~\ref{sec:neural} and~\ref{sec:confidence} address these two issues, respectively.

%
%
%

\subsection{Reducing the Computational Complexity}
\label{sec:neural}

Since $\tilde{f}$, the costs, and the constraints of Eqs.~\ref{eq:ocp} and \ref{eq:structure} are analytic and evaluation time is negligible, the cost of co-design is dominated by $\kappa$. Obtaining a differentiable $\kappa$ from a full simulation with an adjoint method~\cite{Nadarajah00} takes seconds to minutes per call, and the call is repeated at every time step, solver iteration, and design update: about $10^7$ times per co-design run for the settings of the results section. Prior gradient-based methods make this tractable by restricting the flight envelope, the flow model, or the controller until $\kappa$ becomes cheap~\cite{Lee16b,Mackle24}. To avoid this and to handle complex physics in the full envelope, we proceed in two steps. We first introduce a neural surrogate and then approximate it in the inner loop to speed up the computation of the inner problem.

\parag{Neural Surrogate.}
Let $\mathcal{N}_\zeta$ be a network with weights $\zeta$, trained on simulation data so that
\begin{equation}
\hat{\kappa}(\psi, s) = \mathcal{N}_\zeta(\psi, s) \approx \kappa(\psi, s)
\label{eq:surrogate}
\end{equation}
over a large portion of the design space and the full range of operating conditions of the tasks. Replacing $\kappa$ by $\hat{\kappa}$ turns a simulator call into a forward pass and makes Eq.~\ref{eq:ocp} differentiable in $\psi$. However, embedding the network in the solver is still costly since the solver needs $\hat{\kappa}$ and its derivatives at every time step and iteration, with each requiring a pass over all $|\zeta|$ weights, typically millions. With $N$ time steps and $I$ solver iterations, one OCP solve costs $O(N I |\zeta|)$.

\parag{Polynomial Reduction.}
\label{sec:polynomial}
We reduce this cost by exploiting the nested structure of the problem. During an inner solve, the design $\psi$ is fixed, so the surrogate reduces to $\mathcal{N}_\zeta(\psi, \cdot)$, a function of the operating point $s \in \mathbb{R}^{n_s}$ alone, with $n_s$ being small. 
Once per outer iteration, we fit a polynomial to this function over an operating envelope $\mathcal{E} \subset \mathbb{R}^{n_s}$ that covers the conditions the tasks can reach. In the inner loop, the solver then uses the polynomial instead of the network, as described below. 
For $ s \in \mathcal{E}$, we write this as
\begin{equation}
\mathcal{N}_\zeta(\psi, s) \approx \varphi(s)^\top w(\psi) \; .
\label{eq:cheb}
\end{equation}
Here $\varphi(s)$ is the tensor-product Chebyshev basis~\cite{Mason02} with $k$ polynomials $T_0, \dots, T_{k-1}$ per dimension, so $\varphi(s) \in \mathbb{R}^{k^{n_s}}$, and $w(\psi)$ holds one column of coefficients per component of $\kappa$. The fit costs $M$ network evaluations at sample points $s_{1:M} \in \mathcal{E}$. The solver then evaluates $\hat{\kappa}$ and its derivatives analytically at cost $O(|w|)$, so one OCP solve costs $O(M |\zeta| + N I |w|)$ instead of $O(N I |\zeta|)$, with $M \ll N I$ and $|w| \ll |\zeta|$. The inner loop thus no longer depends on the size of the network, and the size of the reduction depends on $n_s$ but not on the design dimension $d$.

We employ Chebyshev polynomials because they are near-optimal in the maximum norm, which matters because the solver may query any point of $\mathcal{E}$, and fitting them is a linear least-squares problem, so the fitting cost is negligible. We place the samples at the Chebyshev nodes~\cite{Brutman78}, which are denser near the ends of each interval to prevent Runge's phenomenon. With Chebyshev nodes, the error decreases steadily as the polynomial degree $k$ increases. We choose $k$ so that the reduction error stays well below the surrogate's own error with respect to the simulator.

\parag{Design Gradient.}
A gradient step on the design requires
\begin{equation}
\frac{d\Phi}{d\psi} = \frac{\partial \Phi}{\partial \psi}
+ \sum_t \frac{\partial \Phi}{\partial x^*_t}\,\frac{d x^*_t}{d\psi}
+ \sum_t \frac{\partial \Phi}{\partial u^*_t}\,\frac{d u^*_t}{d\psi} .
\label{eq:hypergrad}
\end{equation}
Since $x^*$ and $u^*$ are produced by a numerical solver, their derivatives are obtained by applying the implicit function theorem to the optimality conditions of Eq.~\ref{eq:ocp}. The resulting linear system is block-banded and can be solved in $O(N)$ instead of $O(N^3)$~\cite{Xu23g}. The reduction simplifies the rest of the computation since the design affects the inner problem only through the polynomial coefficients $w$, so
\begin{equation}
\frac{dx^*}{d\psi} = \mathrm{D}_w x^* \, \mathrm{D}_\psi w ,
\label{eq:chain}
\end{equation}
where $\mathrm{D}_w x^*$ comes from the implicit differentiation above and $\mathrm{D}_\psi w$ from the fit. The sample points $s_1, \dots, s_M$ are fixed Chebyshev nodes of the envelope $\mathcal{E}$ and do not depend on $\psi$, so the basis matrix of the polynomial $B = [\varphi(s_1), \dots, \varphi(s_M)]^\top$ and its pseudo-inverse $B^{+} = (B^\top B)^{-1} B^\top$ are computed once, and
\begin{equation}
\begin{aligned}
w(\psi) &= B^{+} \, \mathcal{N}_\zeta(\psi, s_{1:M}), \\
\mathrm{D}_\psi w &= B^{+} \, \mathrm{D}_\psi \mathcal{N}_\zeta(\psi, s_{1:M}) .
\end{aligned}
\label{eq:dwdpsi}
\end{equation}
Combining the gradients using the chain rule, the resulting gradient $\nabla_\psi \Phi = \mathrm{D}_\psi \mathcal{N}_\zeta(\psi, s_{1:M})^\top (B^{+})^\top \nabla_w \Phi$ is a single vector-Jacobian product, that is, one backward pass through the network at the $M$ sample points. The network is therefore evaluated once forward and once backward per outer iteration, while the inner solver only uses the polynomial.
 
\subsection{Constraining the Design to the Valid Region}
\label{sec:confidence}

The surrogate is accurate near its training distribution but can degrade away from it, so an
optimizer following its gradients may exploit model errors and find designs with high predicted
but poor true performance~\cite{Trabucco21a,Sharpe25a}. Unlike in surrogate-based shape
optimization, where the model is queried at fixed design conditions, in co-design the
conditions under which the model must be trusted are set by the maneuvers, which change with
the design. Requiring validity everywhere in $\mathcal{E}$ would reject good designs because of
conditions that no maneuver visits, whereas requiring it at preselected conditions could miss
the ones the maneuvers do visit. We therefore require it only where the current optimal
trajectories go.
 
Let $c(\psi, s) \in [0,1]$ be a validity signal that decreases away from the training
distribution, and let $s^*_{j,t} = s(x^*_{j,t}, u^*_{j,t})$ be the operating point of task $j$
at time $t$ along its optimal trajectory. We select the envelope nodes within distance $r$ of
these points,
\begin{equation}
\mathcal{I}(\psi) = \big\{\, i \;:\; \min_{j,t} \| s_i - s^*_{j,t} \| \le r \,\big\},
\label{eq:active}
\end{equation}
with distances measured in the normalized coordinates of $\mathcal{E}$, and impose
\begin{equation}
c(\psi, s_i) \ge c_{\min} \qquad \forall\, i \in \mathcal{I}(\psi)
\label{eq:confidence}
\end{equation}
as one of the design constraints $g(\psi) \le 0$. A design is thus admissible only if the model
can be trusted wherever its own maneuvers take it: the trusted region is not fixed in advance
but follows the trajectories as the design changes.
 
The constraint adds almost no cost. The set $\mathcal{I}(\psi)$ is recomputed once per outer iteration from the solved trajectories and held fixed during differentiation. Thus, the constraint only evaluates $c$ at nodes where the network is already evaluated for the fit, and its gradient comes from the same backward pass, without trajectory sensitivities. The radius $r$ provides a margin for nodes the trajectories may reach after the next design update.
 
The method only requires a scalar validity signal. NeuralFoil~\cite{Sharpe25a} provides one directly; for other surrogates, it can be obtained from ensemble disagreement, conformal prediction intervals, or training-data density.


\subsection{Co-Design Algorithm}
\label{sec:algorithm}
 
We solve the outer problem by gradient descent on the Augmented
Lagrangian~\cite{Hestenes1969multiplier}
\begin{equation}
\begin{aligned}
\mathcal{L}(\psi, \lambda, \nu, \mu) = \Phi(\psi)
&+ \lambda^\top h(\psi) + \nu^\top [g(\psi)]_{+} \\
&+ \tfrac{\mu}{2} \big( \|h(\psi)\|^2 + \|[g(\psi)]_{+}\|^2 \big),
\end{aligned}
\label{eq:auglag}
\end{equation}
where $[\cdot]_{+} = \max(0, \cdot)$, and $g$ collects the geometric requirements, the validity constraint of Eq.~\ref{eq:confidence}, and, with multiple tasks, the requirements $\Phi_j(\psi) - \varepsilon_j$ of Eq.~\ref{eq:multitask}, in which case $\Phi$ is replaced by $\Phi_1$. Algorithm~\ref{alg:codesign} summarizes the procedure.
 
\begin{algorithm}
\caption{Nested co-design with a learned coefficient model}
\label{alg:codesign}
\begin{algorithmic}[1]
\Require design $\psi$, envelope $\mathcal{E}$, surrogate $\mathcal{N}_\zeta$, tasks $1, \dots, J$
\State $s_{1:M} \gets \textsc{ChebyshevNodes}(\mathcal{E})$ \Comment{once}
\Repeat
  \State $\hat{\kappa}_{1:M} \gets \mathcal{N}_\zeta(\psi, s_{1:M})$, \;
         $c_{1:M} \gets c(\psi, s_{1:M})$
  \State $w \gets \textsc{ChebyshevFit}(s_{1:M}, \hat{\kappa}_{1:M})$ \Comment{Eq.~\ref{eq:dwdpsi}}
  \For{$j = 1, \dots, J$ \textbf{in parallel}}
    \State $x^*_j, u^*_j \gets \textsc{SolveOCP}_j(w)$ \Comment{Eq.~\ref{eq:ocp}}
    \State $\nabla_w \Phi_j \gets \textsc{ImplicitDiff}(x^*_j, u^*_j)$ \Comment{\cite{Xu23g}}
  \EndFor
  \State $\mathcal{I} \gets \textsc{ConfidenceNodes}(x^*_{1:J})$ \Comment{Eq.~\ref{eq:active}}
  \State $\nabla_\psi \mathcal{L} \gets \textsc{Backprop}(\mathcal{N}_\zeta, \nabla_w \Phi_{1:J}, c_{\mathcal{I}})$
         \Comment{Eq.~\ref{eq:dwdpsi}, \ref{eq:auglag}}
  \State $\psi \gets \textsc{Adam}(\psi, \nabla_\psi \mathcal{L})$; update $\lambda, \nu, \mu$
         \Comment{Eq.~\ref{eq:auglag}}
\Until{convergence}
\end{algorithmic}
\end{algorithm}

\parag{Properties.}
The sensitivities are exact for the reduced problem, up to the inner solver's tolerance, so each step follows a descent direction of the model rather than a finite-difference or sampled estimate of one. The cost of an outer iteration is linear in the horizon $N$ by construction of
the implicit differentiation, its inner loop does not depend on the network size, and it grows at most linearly with the design dimension $d$, which enters only through one vector-Jacobian product. As shown in the results section, this is what makes high-dimensional shape parameterizations tractable.


\begin{figure*}[h]
    \centering
    \subfloat[\label{fig:glider-sketch}]{%
        \definecolor{orE}{HTML}{F16913}\definecolor{orI}{HTML}{7F2704}
\definecolor{bodyCol}{HTML}{3A3A3A}\definecolor{skinCol}{HTML}{ECECEC}\definecolor{elevCol}{HTML}{BFBFBF}
\definecolor{refCol}{HTML}{A6A6A6}
\resizebox{\columnwidth}{!}{%
\begin{tikzpicture}[x=1cm, y=1cm, font=\small,
  vec/.style={-{Stealth[length=4pt,width=3pt]}, line width=0.6pt},
  force/.style={vec, preaction={draw=white, line width=1.8pt, -}},
  ref/.style={refCol, dash pattern=on 2pt off 1.5pt, line width=0.35pt},
  axisline/.style={bodyCol!55, dash pattern=on 2.5pt off 1.5pt, line width=0.4pt},
  ext/.style={refCol, line width=0.3pt},
  dim/.style={{Stealth[length=2.6pt,width=2.2pt]}-{Stealth[length=2.6pt,width=2.2pt]},
              line width=0.35pt, bodyCol},
  dlab/.style={font=\footnotesize, inner sep=1pt},
  skin/.style={fill=skinCol, draw=bodyCol, line width=0.55pt, line join=round},
  com/.style={circle, fill=bodyCol, draw=white, line width=0.4pt, inner sep=0pt, minimum size=4pt},
  declare function={
    xc(\q)=(1-cos(\q))/2;
    yt(\x)=5*0.10*(0.2969*sqrt(\x)-0.1260*\x-0.3516*\x^2+0.2843*\x^3-0.1015*\x^4);
    yc(\x)=(\x<0.4) ? 0.03/(0.16)*(0.8*\x-\x*\x) : 0.03/(0.36)*(0.2+0.8*\x-\x*\x);
  }]
\draw[ref] (-2.3,0) -- (1.9,0);                                    
\begin{scope}[rotate=12]
  \coordinate (O) at (0,0);                                         
  \coordinate (A) at (-0.33,0);                                     
  \coordinate (W) at (-0.69,0.008);                                 
  \coordinate (F) at (-1.25,-0.10);                                 
  \coordinate (H) at (-3.7,0);                                      
  \coordinate (E) at ($(H)+(200:0.45)$);                            
  \draw[ext] (O) -- (0,-0.84);
  \draw[ext] (W) -- (-0.69,-0.42);
  \draw[ext] (F) -- (-1.25,-0.63);
  \draw[ext] (H) -- (-3.7,-0.84);
  \draw[dim] (0,-0.36) -- (-0.69,-0.36);
  \node[dlab, anchor=east] at (-0.73,-0.36) {$\ell_w$};
  \draw[dim] (0,-0.57) -- (-1.25,-0.57);
  \node[dlab, anchor=east] at (-1.29,-0.57) {$\ell_f$};
  \draw[dim] (0,-0.78) -- (-3.7,-0.78);
  \node[dlab, anchor=north] at (-1.85,-0.80) {$\ell$};
  \draw[ext] (H) -- ++(110:0.34);
  \draw[ext] (E) -- ++(110:0.34);
  \draw[dim] ($(H)+(110:0.28)$) -- ($(E)+(110:0.28)$);
  \node[dlab, anchor=south] at ($(H)+(200:0.22)+(110:0.35)$) {$\ell_e$};
  \draw[ref] (1.3,0) -- (1.78,0);                                   
  \draw[ref] (H) -- ++(180:0.85);                                   
  \draw[skin, rounded corners=1.2pt]                                
    (-3.10,0.03) -- (-3.50,0.66) -- (-3.70,0.66) -- (-3.72,0.03) -- cycle;
  \draw[skin]                                                       
    (1.30,0.00) .. controls (1.26,0.12) and (0.80,0.23) .. (0.20,0.25)
    .. controls (-0.60,0.27) and (-1.60,0.22) .. (-2.30,0.10)
    .. controls (-2.60,0.05) and (-2.75,0.04) .. (-2.95,0.04)
    -- (-3.72,0.035) .. controls (-3.77,0.035) and (-3.77,-0.03) .. (-3.72,-0.03)
    -- (-2.95,-0.035)
    .. controls (-2.65,-0.04) and (-2.45,-0.07) .. (-2.00,-0.12)
    .. controls (-1.40,-0.17) and (0.40,-0.18) .. (0.90,-0.11)
    .. controls (1.15,-0.08) and (1.30,-0.04) .. (1.30,0.00) -- cycle;
  \draw[axisline] (H) -- (1.3,0);                                   
  \begin{scope}[shift={(H)}, rotate=200]                            
    \draw[skin, fill=elevCol] (0,0.03) .. controls (0.12,0.045) and (0.4,0.03) .. (0.85,0.0)
      .. controls (0.4,-0.03) and (0.12,-0.045) .. (0,-0.03)
      .. controls (-0.025,-0.02) and (-0.025,0.02) .. (0,0.03) -- cycle;
  \end{scope}
  \draw[line width=0.45pt] (H) ++(180:0.6) arc (180:200:0.6);
  \node at ($(H)+(190:0.82)$) {$\phi$};
  \begin{scope}[shift={(0.12,-0.046)}, xscale=-1.8, yscale=1.8]     
    \draw[fill=orI, draw=orI!70!black, line width=0.4pt]
      plot[domain=0:180, samples=60, variable=\q] ({xc(\q)},{yc(xc(\q))+yt(xc(\q))})
      -- plot[domain=180:0, samples=60, variable=\q] ({xc(\q)},{yc(xc(\q))-yt(xc(\q))}) -- cycle;
  \end{scope}
  \node[com] at (W) {};
  \node[com] at (F) {};
  \node[com] at (E) {};
\end{scope}
\fill[white, draw=bodyCol, line width=0.5pt] (0,0) circle (2.2pt);  
\node[inner sep=1pt, anchor=west] at (0.13,0.05) {$O$};
\draw[line width=0.45pt] (180:1.65) arc (180:192:1.65);
\node at (186:1.9) {$\theta$};
\coordinate (Ag) at (192:0.33);
\draw[vec, draw=orE] (Ag) -- ++(-10:2.05) node[right, text=orE] {$-\bm{v}_w$};
\draw[line width=0.45pt] ($(Ag)+(12:1.85)$) arc (12:-10:1.85);
\node at ($(Ag)+(1:2.07)$) {$\alpha$};
\draw[force, draw=bodyCol] (Ag) -- ++(80:1.0) node[above] {$F^L_w$};
\draw[force, draw=bodyCol] (Ag) -- ++(170:0.8) node[above left, inner sep=1pt] {$F^D_w$};
\end{tikzpicture}}}
    \hfil
    \subfloat[\label{fig:glider-params}]{%

\begin{footnotesize}
\begin{tabular}{l c c}
\hline
Parameter & Symbol & Value \\
\hline
Fuselage mass & $m_f$ & 0.026 $kg$ \\
Wing mass & $m_w$ & 0.035 $kg$ \\
Elevator mass & $m_e$ & 0.004 $kg$ \\
Distance to elevator hinge & $\ell$ & 0.260 $m$ \\
Elevator COM position & $\ell_e$ & 0.020 $m$ \\
Fuselage COM position & $\ell_f$ & -0.027 $m$ \\
Wing COM position & $\ell_w$ & -0.010 $m$ \\
Wing Surface Area & $S_w$ & 0.158 $m^2$ \\
Elevator Surface Area & $S_e$ & 0.017 $m^2$ \\
\hline
\end{tabular}
\end{footnotesize}
}
    \vspace{-2mm}
    \caption{\small \textbf{Glider model.} (a) Dots mark the centers of mass. Lengths on the
             left are measured from $O$. (b) Model parameters.}
    \label{fig:glider}
\end{figure*}

\section{Experimental Evaluation}

We co-design the wing of the small glider in Fig.~\ref{fig:glider-sketch} together with its control inputs; the rest of the glider is fixed. We consider two cases: a \emph{2D wing}, defined by a single cross section, and a \emph{3D wing}, defined by several cross sections along the span. The glider must perform three tasks of increasing difficulty: flying as far as possible, landing in the shortest distance, and perching at a given location at near-zero speed (Fig.~\ref{fig:tasks}). These tasks pull the wing in opposite directions: gliding favors thin, low-drag airfoils, short landing favors sections that dissipate energy quickly, and perching requires decelerating through stall, where the aerodynamics are hardest to model. After describing the model and the tasks (Secs.~\ref{sec:model} and~\ref{sec:exp-tasks}), we use the 2D wing to show that our approach outperforms the alternatives and can trace the Pareto front of conflicting tasks (Sec.~\ref{sec:exp-airfoil}), the 3D wing to show that it scales linearly with the number of parameters (Sec.~\ref{sec:exp-wing3d}), and ablations to isolate the reduction and the validity constraint (Sec.~\ref{sec:exp-ablation}).

\subsection{Glider Model}
\label{sec:model}

\parag{Dynamics}.
The state of the glider is
\begin{equation}
\bm{x} = \begin{bmatrix} x & z & \theta & \phi & \dot{x} & \dot{z} & \dot{\theta}\end{bmatrix}^\top ,
\label{eq:state}
\end{equation}
where $(x,z)$ is the center-of-mass position, $\theta$ the pitch angle, and $\phi$ the elevator angle; the control is $u = \dot\phi$. We use the planar rigid-body model of~\cite{Moore14} with the parameters of Fig.~\ref{fig:glider-params}. In the notation of Eq.~\ref{eq:structure}, the design-dependent coefficients are the wing's lift, drag, and moment coefficients, $\kappa = (C_L, C_D, C_M)$, and the operating point is $s = (\alpha, Re)$: the wing's angle of attack and Reynolds number $Re = \|\bm{v}_w\| \bar{c}_w / \nu$, where $\bm{v}_w$ is the air velocity relative to the wing.
The coefficients are used to define the aerodynamic forces and moments acting on the wing as
\begin{equation}
\begin{aligned}
F^L_w &= \tfrac{1}{2}\rho \|\bm{v}_w\|^2 S_w\, C_L(\psi, s), \\
F^D_w &= \tfrac{1}{2}\rho \|\bm{v}_w\|^2 S_w\, C_D(\psi, s), \\
M_w   &= \tfrac{1}{2}\rho \|\bm{v}_w\|^2 S_w \bar{c}_w\, C_M(\psi, s).
\end{aligned}
\label{eq:aeroforces}
\end{equation}
The forces and moments then specify the dynamics $\tilde{f}$ of Eq.~\ref{eq:structure} through the rigid-body
equations of motion
\begin{equation}
\begin{aligned}
m \begin{bmatrix}\ddot{x} \\ \ddot{z}\end{bmatrix} &= \bm{F}_w + \bm{F}_e
  - \begin{bmatrix} 0 \\ m g \end{bmatrix}, \\
I \ddot{\theta} &= \bm{r}_w \times \bm{F}_w + \bm{r}_e \times \bm{F}_e + M_w + M_e,
\end{aligned}
\label{eq:eom}
\end{equation}
together with $\dot\phi = u$. Here $\bm{F}_w$ is the wing force, with component $F^D_w$ along $\bm{v}_w$ and $F^L_w$ normal to it; $\bm{F}_e$ and $M_e$ are the elevator force and moment, computed in the same way but with coefficients that do not depend on the design; $\bm{r}_w$ and $\bm{r}_e$ point from the center of mass to the aerodynamic centers of the wing and elevator; and $m$ and $I$ are the total mass and pitch inertia. We use 4th-order Runge-Kutta \cite{Butcher16} to integrate Eq.~\ref{eq:eom}, yielding the value of $f$ in Eq.~\ref{eq:structure}.

\parag{Shape Parameterization.}
Each wing section is a Kulfan (CST) airfoil~\cite{Kulfan08} described by 18 parameters: 8 upper and 8 lower weights, a leading-edge weight, and the trailing-edge thickness. The \emph{2D wing} uses a single section along the whole span, so $\psi \in \mathbb{R}^{18}$. The \emph{3D wing} uses $N = 1, \dots, 7$ independent sections at fixed spanwise stations per wing, interpolated between stations, so $\psi \in \mathbb{R}^{18N}$, yielding from 18 to 126 parameters. A minimum thickness of 5\% of the chord holds at every chordwise station of every section.

\parag{Aerodynamic Model and its Reduction.}
For a 2D wing,  we use NeuralFoil~\cite{Sharpe25a}, a network trained on XFoil~\cite{Drela89} simulations, to compute $\kappa$. It is differentiable in $\psi$, about three orders of magnitude faster than XFoil, defined through post-stall, and it returns a confidence $c(\psi, \alpha, Re)$.  
For a 3D wing, a lifting line~\cite{Phillips00}  combines NeuralFoil values over the $N$ spanwise panels, each at its local angle of attack and Reynolds number. In both cases, $\kappa$ only depends on $s = (\alpha, Re)$ and the reduction of Sec.~\ref{sec:polynomial} applies similarly. At each design update, the wing model is evaluated on a $32\times32$ grid of Chebyshev nodes over the flight envelope $\alpha \in [-12^\circ, 23^\circ]$, $\log_{10} Re \in [\log_{10} 3500, \log_{10} 45000]$, and reduced to a $26\times26$ Chebyshev expansion.


\subsection{Tasks and Protocol}
\label{sec:exp-tasks}
\parag{Tasks.}
Each task is an OCP of the form of Eq.~\ref{eq:ocp} with a free time step~\cite{Moore14}, solved from $K = 4$ launch states; $\Phi(\psi)$ is the mean cost over them, so that the wing must work across a range of launch conditions instead of being tailored to one. For gliding and short landing, the glider is released horizontally at $\dot x_0 = 6$\,m/s from $x_0 = 0$ at four heights, $z_0 \in \{2,3,4,5\}$\,m, so that the energy it must shed before touchdown nearly doubles across launches. For perching, it is launched upward at $(\dot x_0, \dot z_0) = (6, 3)$\,m/s from 10--11\,m before the perch and 0--1\,m above it ($x_0 \in \{-11,-10\}$\,m, $z_0 \in \{0,1\}$\,m), far enough that the initial design cannot reach the perch at near-zero speed. All other state components start at zero, and the horizons are $N = 100$ for perching and landing and $N = 200$ for gliding. Every task enforces the actuator limits $-\pi/3 \le \phi \le \pi/8$ and $|\dot\phi| \le 13$\,rad/s, the minimum-thickness constraint, and the validity constraint of Eq.~\ref{eq:confidence} with $c_{\min} = 0.6$ and $r = 0.1$ in the normalized coordinates of $\mathcal{E}$. 
\emph{Gliding} must maximize the horizontal distance before a soft touchdown.
\emph{Short landing} must touch down softly as close as possible, with a terminal cost on position, sink rate, and pitch. 
\emph{Perching} must reach the perch pose at near-zero velocity.
 
\parag{Baselines and Protocol.}
The baselines cover the alternatives currently available to a practitioner: designing the wing first, searching over wings without using gradients, and optimizing wing and trajectories in a single problem.
\emph{Initial design} flies the initial airfoil optimally. \emph{Sequential}~\cite{Allison14} first optimizes a task-agnostic aerodynamic proxy over the envelope (maximum $L/D$, maximum $C_{L,\max}$, or maximum high-$\alpha$ drag), then flies the result optimally. GA~\cite{Bergonti24}, CMA-ES~\cite{Hansen16}, and TuRBO~\cite{Eriksson19a}
perform nested black-box co-design: an outer search over the design whose candidates are evaluated by the same inner OCP. \emph{Simultaneous}~\cite{Spielberg17} optimizes $\psi$ and all trajectories in a single NLP, with an exact Hessian or L-BFGS.  All methods use the same validity rule, inner OCP, constraints, and initial designs (NACA 4412, NACA 0012, Clark Y), with a 4\,h wall-clock budget on 4 CPU cores. Every final design is re-scored by re-solving its OCPs with XFoil tables in place of the model.

\subsection{2D Wing: Comparison and Multi-Objective Optimization}
\label{sec:exp-airfoil}


\begin{figure}
    \centering
    \includegraphics[width=\linewidth]{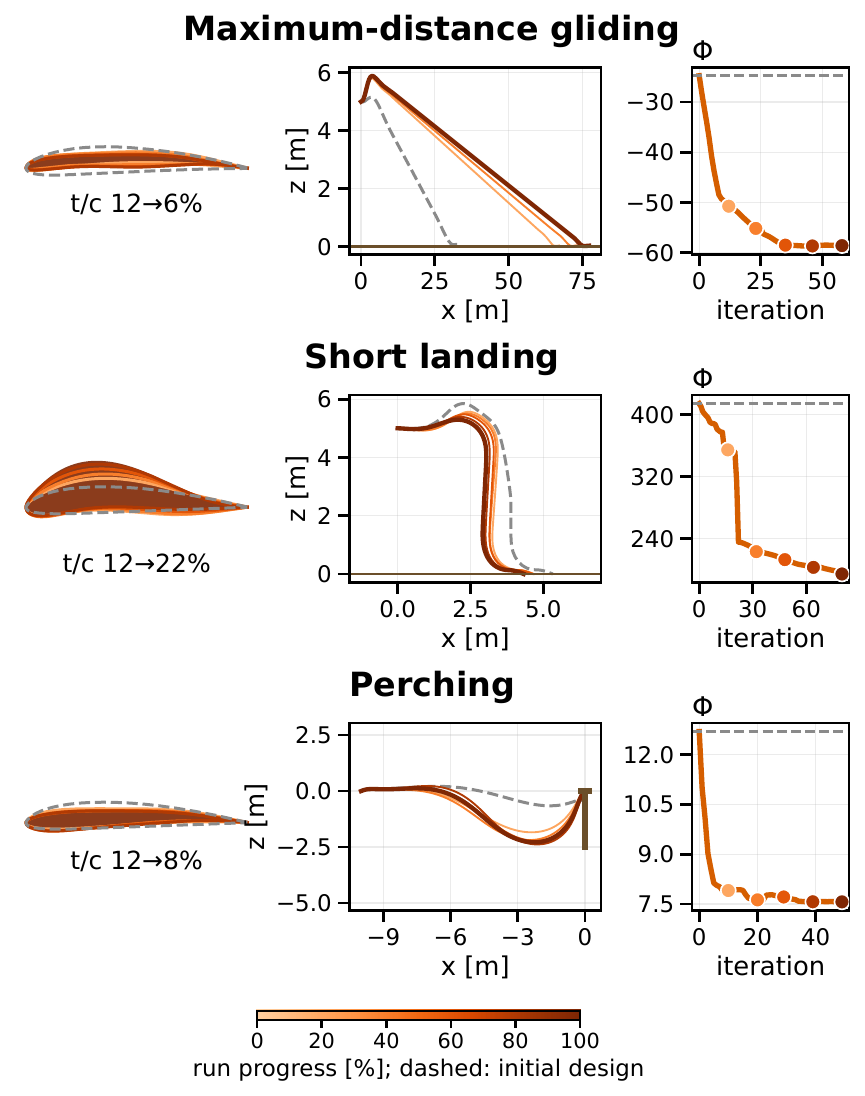}
    \vspace{-6mm}
    \caption{\small \textbf{Tasks and controls.} Optimized airfoils (left), trajectories (middle), and mission cost over the outer iterations (right) for the three tasks, colored by run progress (dashed: initial design).}
    \label{fig:tasks}
\end{figure}

\begin{table}[h]
\begin{tabular}{@{}lrrrr@{}}
\toprule
Method & Gliding & Landing & Perching & min \\
\midrule
Initial design & $-$24.3$\pm$0.5 & -- (0/3) & -- (2/3) & 16 \\
Seq.\ max L/D & $-$40.8$\pm$12.8 & -- (2/3) & -- (1/3) & 93 \\
Seq.\ max $C_{L,\max}$ & $-$25.5$\pm$4.9 & -- (2/3) & 25.49$\pm$14.79 & 83 \\
Seq.\ max $C_D$ & $-$37.3$\pm$2.0 & -- (2/3) & -- (2/3) & 81 \\
GA & $-$33.0$\pm$7.2 & 244$\pm$7 & 8.63$\pm$0.19 & 246 \\
CMA-ES & $-$39.3$\pm$9.5 & -- (2/3) & 9.13$\pm$0.57 & 265 \\
TuRBO & $-$37.3$\pm$6.2 & 230$\pm$15 & -- (2/3) & 261 \\
Simult.\ (exact H) & $-$26.2$\pm$3.1 & 248$\pm$6 & -- (2/3) & 281 \\
Simult.\ (L-BFGS) & -- (1/3) & -- (1/3) & -- (0/3) & 279 \\
Ours & \textbf{$-$60.2$\pm$2.1} & \textbf{184$\pm$28} & \textbf{8.42$\pm$0.37} & 47 \\
\bottomrule
\end{tabular}
\caption{\textbf{Comparison with the baselines.} Mission cost $\Phi$ (lower is better), re-scored with XFoil. Mean $\pm$ std over I1--I3 if feasible on all three, else the number of feasible inits ($t/c \ge 0.05$, $c \ge 0.6$ at visited operating points). \textbf{Bold}: best; min: median wall-clock.}
\label{tab:main}
\end{table}

We first demonstrate that our approach to nested co-design with gradient outperforms designing shape and controls separately or searching over wings without using gradients. We then show that our approach enables us to effectively sample the Pareto front of possible solutions.  

\parag{Comparative Results.}
In Tab.~\ref{tab:main}, we report comparative results against the baselines. Among the methods feasible from all three initial designs, ours reaches the lowest cost on every task, with a median wall-clock of 47\,min against the 4\,h budget of the other methods. The baselines fall short for three reasons. Black-box search runs out of evaluations: GA, CMA-ES, and TuRBO evaluate only 11--79 designs in 4\,h, because each new design is solved far from any warm start, with single solves taking up to 52\,min, whereas ours follows the gradient and warm-starts every solve from the previous iterate, at 29\,s per iteration. The simultaneous NLP is too large and nonconvex: it hits the time limit in all 12 runs, 6 of them at infeasible points. Finally, aerodynamic proxies help only when they happen to match the task: maximizing $L/D$ is the second-best method for gliding, yet its wings glide 32\% less far than ours (41 vs.\ 60\,m on average), and no proxy yields a competitive perching or landing design.
 
Fig.~\ref{fig:tasks} shows that each task drives the airfoil in a different direction. For perching, the airfoil thins ($t/c$ from 11.7\% to 8.5\%), and the glider dives deeper before pulling up onto the perch ($z \approx -2.3$\,m instead of $-0.6$\,m). For short landing, the thickness nearly doubles (21.7\%) and the camber increases, so that the wing acts as an air brake during the near-vertical descent. For gliding, the airfoil becomes thin and cambered ($t/c$ 6.2\%), as is typical of high lift-to-drag airfoils at low Reynolds numbers~\cite{Lissaman83,Mueller23a}, and the mean glide distance grows from 24 to 60\,m.


\begin{figure}[h]
    \centering
    \includegraphics[width=\linewidth]{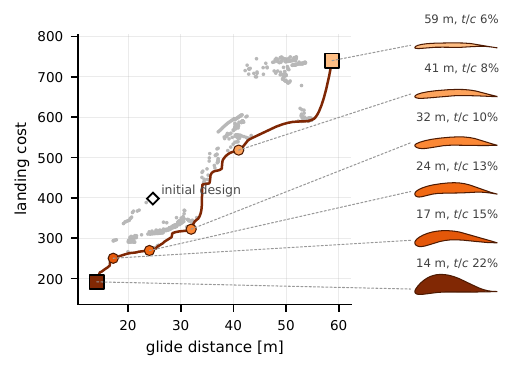}
    \vspace{-8mm}
    \caption{\small \textbf{Pareto front between landing cost and glide distance.} The line interpolates all the non-dominated designs among all those evaluated by the sweep. Gray represents dominated designs.}
    \label{fig:pareto-front}
\end{figure}

\parag{Multi-Objective Optimization.}
A robot is rarely built for a single maneuver. We use the $\varepsilon$-constraint formulation of Eq.~\ref{eq:multitask} to trace the trade-off between short landing distance and gliding ability: we minimize the landing cost subject to a minimum glide distance $\varepsilon$, and sweep $\varepsilon$ to produce the Pareto front of Fig.~\ref{fig:pareto-front}. The 147 non-dominated designs, among the 730 evaluated, are ordered by their geometry: thickness and camber decrease almost monotonically with glide distance (Spearman correlations $-0.99$ and $-0.96$), from the thick, cambered landing section to the thin gliding one. The front is steep and uneven: raising the glide distance from 14 to 21\,m increases the landing cost from 192 to 265, which then stays nearly flat up to 32\,m before rising to 473 at 37\,m. This non-convex shape is what the $\varepsilon$-constraint formulation can trace, unlike a weighted sum of the two costs~\cite{Das97}.

\subsection{3D Wing: Scaling the Design Dimension}
\label{sec:exp-wing3d}


\begin{figure}[h!]
    \centering
    \includegraphics[width=\linewidth]{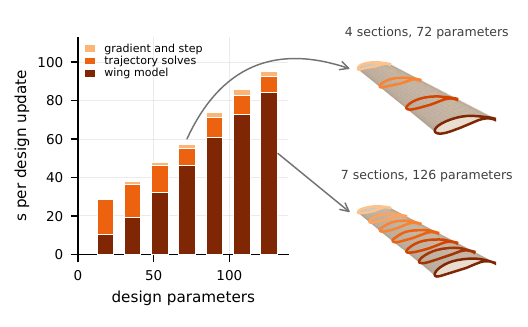}
    \vspace{-8mm}
    \caption{\small \textbf{3D wing, perching task.} Time per design update versus number of design parameters. Only the time to compute the wing model grows. The insets show wings with 4 and 7 sections.}
    \label{fig:3dwing}
\end{figure}

Current Co-Design methods \cite{Bergonti24, Fardini21} use 10 to 20 parameters. To show that our approach can handle far more, we now switch to the 3D wing, with $N$ separate sections, for $N = 1, \dots, 7$, which yields from 18 to 126 parameters. We report the results in Fig.~\ref{fig:3dwing} for perching from a single launch state. The time of a design update increases linearly, from 29 to 95\,s, and the growth (11 to 84\,s) comes entirely from the increased number of NeuralFoil evaluations in line with the number of lifting-line panels. The cost of solving for trajectory remains constant, and evaluating the gradient, the only step that involves the design parameters, takes a few seconds at every dimension, because the sensitivities of all parameters come from a single vector--Jacobian product. 

\subsection{Ablations}
\label{sec:exp-ablation}
%

\begin{figure}[h!]
    \centering
    \includegraphics{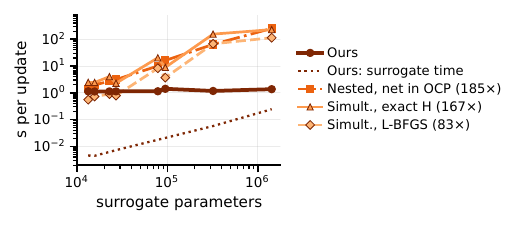}
    \vspace{-8mm}
    \caption{\small \textbf{Time per design update versus surrogate size.} Ours stays constant; with the network inside the OCP or the simultaneous NLP, the cost grows.}
    \label{fig:scaling}
\end{figure}

\parag{Polynomial Reduction Pays off.}
Here we show that the polynomial reduction introduced in Sec.~\ref{sec:neural} allows us to scale up the surrogate size.
In Fig.~\ref{fig:scaling}, we plot the computation time of our method against actually evaluating the surrogate model inside the inner loop, or inside the simultaneous NLP, for several sizes of the surrogate model, on a single perching OCP from one launch state. 
Decoupling the inner problem from the network as we do makes our update independent of the surrogate size: about 1.4\,s from the smallest to the largest network, of which network evaluation takes at most 0.08\,s. With the network in the inner loop, every solver iteration evaluates it at every node, and at the size we use, a single inner solve or NLP iteration is already 83--185$\times$ slower than our entire update. The speed-up does not cost accuracy: the reduction changes the mission cost by only 0.03--0.39\% with respect to NeuralFoil inside the OCP, across 4 designs.
%

\begin{figure}[h]
    \centering
    \includegraphics[width=\linewidth]{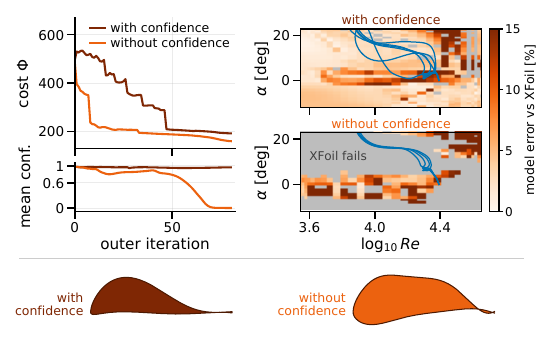}
    \caption{\textbf{Without the confidence constraint, the optimizer exploits the model.} Short landing. Left: model cost and mean confidence per iteration. Right: model error against XFoil for each final design (gray: XFoil fails; blue:
flights). Bottom: final airfoils.}
    \label{fig:no-conf}
\end{figure}

\parag{Validity of the Final Design.}
The confidence constraint of Sec.~\ref{sec:confidence} prevents optimizer exploitation of the learned model. Fig.~\ref{fig:no-conf} compares short landing results with and without this constraint. Without it, the predicted cost is 17\% lower (160 vs.\ 192), but surrogate confidence drops to nearly zero. The resulting 33\%-thick section exhibits an artifact, with the upper curve crossing the lower one, and cannot be reliably validated: XFoil fails to converge at 92\% of operating points near the flight trajectory, with a median model error of 13\% at the remaining points. With the constraint, confidence remains at 0.97 and the median model error is only 3.5\%.


\section{Conclusion}

In this paper, we presented a scalable bilevel co-design framework that jointly optimizes aerodynamic shape and motion planning. By combining a gradient-based Augmented Lagrangian upper optimizer with differentiable optimal control and a learned aerodynamic surrogate, our approach optimizes continuous, high-dimensional design spaces that are intractable for sampling-based methods. 

While the current work focuses on aerodynamic design and planar trajectories, the framework is completely general and designed to scale to more complex problems. A natural extension is to full 3D trajectory optimization. More broadly, the bilevel structure of our pipeline is compatible with mixed-integer optimization, which would allow the framework to jointly optimize both individual component shapes and the overall system topology, opening the door to automated structural design of aerial robots.

\bibliographystyle{IEEEtran}

\bibliography{bib/string,bib/cfd,bib/learning,bib/robotics,bib/optim,bib/vision,bib/misc,bib/new}

\end{document}